%% file: paper.tex
\documentclass[letterpaper, 10 pt, conference]{ieeeconf}
\IEEEoverridecommandlockouts
\usepackage{todonotes}

\title{\LARGE \bf 
  A Critical Look at Smart Wheelchairs % Not sure I like this title; change it if you want, but try to keep it neutral
}

\author{Benjamin Narin$^{1}$, Makenzie Brian$^{1}$, and William D. Smart$^{1}$% <-this % stops a space
\thanks{*This work was supported, in part, by the National Science Foundation, under award 1541251.}% <-this % stops a space
\thanks{$^{1}$The authors are with the Collaborative Robotics and Intelligent Systems (CoRIS) Institute, Oregon State University, Corvallis, OR 97331, USA.\ \ %
        {\tt\small narinb@oregonstate.edu}, {\tt\small brianm@oregonstate.edu}, {\tt\small smartw@oregonstate.edu}}%
}

\begin{document}

\maketitle
\thispagestyle{empty}
\pagestyle{empty}

%\begin{abstract}
  \input{abstract}

%\end{abstract}

\input{intro}
\input{related}
\input{complications}

\input{conclusions}
% \input{acknowledgements}

\bibliographystyle{IEEEtran}
\bibliography{paper}

% Remove this once we're done
% \listoftodos

%\addtolength{\textheight}{-12cm}

\end{document}

%% file: abstract.tex
\begin{abstract}Research into smart wheelchairs has been conducted for decades, but we have yet to see the widespread use of this technology among full-time wheelchair users. We argue that the main reason for this is that there is a mismatch between research and the actualities of using a powered mobility device in the real world. Based on our own research experiences, we enumerate some of these disparities, and offer some suggestions for where work in smart wheelchairs might focus in the coming years.
\end{abstract}
%Wheelchairs have long been used as a platform for navigation and localization. In this paper we will argue that wheelchairs cannot be treated as robots for practical intelligent wheelchairs and should be pursued in a variety of other ways. 

%% file: intro.tex
\section{INTRODUCTION}
\label{sec:intro}

Research into smart wheelchairs has been carried out since the 1980s~\cite{Krogh:earlywork,simpson:wheelchair}. However, after almost forty years of effort, we have yet to see a widespread adoption of this technology. Although some self-driving wheelchairs are actually operating in the real world~\cite{spectrum:wheelchair}, they are designed to carry people who are not normally powered wheelchair users. We believe this is a key point that greatly simplifies the problem. \textit{Wheelchairs with sensors are simply not the same as robots.}

%Much of the existing work in smart wheelchairs has focused on navigation and localization, and has treated the chair as, essentially, a mobile robot.  It has sensors, computation, and can be driven around, just like a traditional mobile robot can.  This has allowed researchers to take the large body of work in robotics, and apply it to smart wheelchair research.  However, much of the work reported in the literature is carried out without a person in the chair.  Almost none of the work\todo{Verify this claim, and remove ``almost'' if we can} that we are aware of is done with a full-time wheelchair user.  As we discuss in Section~\ref{sec:occlusion}, people block the sensors mounted on the chair, hang personal belongings on the chair, and change the range of acceptable navigation motions that can be performed.

Despite the fact that there is a large body of excellent work in this area, we claim that there is still a long way  to go before we can realize the full potential of smart wheelchairs. Critically, we must take a hard look at the realities of powered mobility devices as they are used in everyday life.

We begin with a brief survey of related work, offer thoughts on the complications of designing a practical smart wheelchair system for everyday use, and finish by offering some suggestions for future directions in this field.

%Since the early 1980's, intelligent wheelchairs have been a topic of research. In that time, much of the work done has been in navigation and localization, disregarding the requirements of an actual wheelchair which in many cases require instrumentation that would hinder or completely block a wheelchair user from the most mundane use of the chair. In the past, intelligent wheelchair research has been a good sell for grants, publicity, and fulfillment. But if we are to expect practical applications to come from the research we need to take a hard look at what the requirements are for a real system and the complications that arise. The assumption that a powered wheelchair + sensors = a robot does not apply.\\

%By no means are we saying that fully autonomous wheelchairs cannot be done. Only that the problem is much more complex than addressed in the literature. If real world deployment is a goal, there are other better ways to accomplish real world intelligent wheelchairs.

%% file: related.tex
\section{RELATED WORK}
\label{sec:related}

Smart wheelchairs incorporate sensors that allow the system to perceive its environment and react to it. Leaman and La~\cite{leaman:survey} give a comprehensive survey of smart wheelchair research. While there are many such systems described in the survey, we have noticed some commonalities.

Many of the wheelchairs that they describe are used in a research setting, often at a university. A number of these systems, 48 of the 155 cited papers, focus on a navigation task where a human user is not required to be considered. While navigation is important, it also fails to consider complexities of the real world.

%MOVED While sensors are obviously necessary, and more data is always better, some systems~\cite{Kobayashi:sensors} incorporate a large number of costly sensors. This is useful for research purposes, but creates problems when transitioning to a production setting where cost must be minimized.

Additionally, a common theme in smart wheelchair work is to not address long term applications. This assumption is indicative of the lack of real-world focus currently in the area of smart wheelchairs.

%A number of the more recent papers (for example \todo[inline]{cite them here}) compare smart wheelchairs to self-driving cars.  While the analogy is appealing, it is incomplete and, we argue in Section~\ref{sec:complications}, can be misleading.

% Try not to pick on any particular work here, since we don't want to offend any one group of people.

%Lit review showing that most smart wheelchairs have been used in research and do not take into account the user. Those which do have limited applications. Show how the most recent lit review talks about the future of self driving etc, without actually addressing the issues that arise. Include expensive over the top systems (MIT).

%% file: complications.tex
\section{Complications}
\label{sec:complications}

In this section, we discuss the complications of implementing a smart wheelchair system for practical, daily use.

\subsection{Physical Structure and Environment}

It is tempting to analogize a smart wheelchair to a robot or a self-driving car~\cite{leaman:survey}, since these share the same basic components: sensing, computation, and mobility. But this analogy breaks down in a subtle way: for a robot or a car, we can create a physical model, used by the navigation system, that does not change over time. We know, with great accuracy, where the physical boundaries of the robot are. While we can do this for the wheelchair itself, it will be different when there is someone using it. People will move the seat, hang things on the side of the chair, and lean outside the defined boundary.  All of these things change the ``shape'' of the wheelchair, and the volume that we must check for collisions.

\subsection{Environment and Social Conventions}

By simply keeping the wheelchair further away from obstacles, we can address these un-modeled elements. This works in a building that complies with the Americans with Disabilities Act (ADA). However, due to cost, many full-time wheelchair users live in ADA non-compliant spaces, with narrow hallways and doors. Padding the navigation system with a safety buffer can cause problems in these places, since doorways might be too confined for the system to safely plan a path through.

%To build a robust autonomous system, structure is required to allow models to plan actions or figure out what is going on. Often self driving wheelchairs are equated with self driving cars or mobile robots with the assumption that the challenges and constraints are similar, but this is not the case.\\

Self-driving cars work, in part, because of an assumption that there is a highly detailed map in which the car is well-localized~\cite{jo:autonomouscars}. While we can build maps for our smart wheelchairs, they are subject to drastic change over time, as rooms are rearranged, and people wander through. Localization, too, is problematic in cluttered environments. A more serious inconsistency in the analogy with self-driving cars is the assumption of a set of fixed social rules. Cars drive on a particular side of the road and usually follow well-defined protocols. This is not true of wheelchairs, which must operate in a more fluid social environment. How close they can come to people and objects varies by context and cultural setting. While the repercussions of navigation mistakes in a smart wheelchair are less likely to be fatal than in an automobile, they are far easier to make, and can lead to confusion, embarrassment, and physical harm.

\subsection{Sensors and Control}
\label{sec:occlusions}

Individuals modify their wheelchairs, both to personalize them and to install medically-necessary equipment on them. Backpacks, catheter bags, computers, respirators, and other equipment adorn the chairs of most long-term users. These things are tricky to model, and can obscure sensors on the chair. The user's feet, which typically are in front of the chair, occlude one of the most critical regions for navigation. This leaves us with one option: mount the sensors above or outside of the areas where they could be blocked. However, this can lead to a larger chair footprint, and odd-looking ``sensor masts'' which, anecdotally, are aesthetically unappealing to many of the wheelchair users we have spoken with.

In order to use sensors, we must know where they are in respect to some fixed coordinate frame, often the base of the robot.  If we mount them on a moving part of the robot, such as an arm, then we must have sensors that accurately report the position of that part, so that we can calculate where the sensor is. In many powered wheelchairs, the seat can be moved by the user to a variety of positions. Mounting the sensor on a part of the seat, such as the footrest, is similar to mounting it on the arm. However, commercial wheelchairs typically do not allow access to or possess internal position encoders, so determining the sensor position can be impossible. A manufacturer could enable this ability, but there is little commercial incentive to do so. The same is true of the drive system; a direct programmatic interface to the wheel motors and the ability to read the wheel rotation sensors would greatly simplify work with smart wheelchairs.

While sensors are clearly necessary, and more data is always better, some systems~\cite{Kobayashi:sensors} incorporate a large number of costly sensors. This is useful for research purposes, but creates problems when transitioning to a production setting where cost must be minimized.

\subsection{Cost and Medical Insurance}

Additionally, devices that are not ``medically necessary'' are often not covered by insurance in many countries. Much of the functionality of smart wheelchairs is not ``necessary'' by the technical definition. This leads to the question of who pays for this functionality and what level of cost is realistic.

\subsection{Risk}

% Assuming everything previously worked out the device would still be considered a medical device. Although there is legislation currently being worked on for self driving cars there is no such legislation for self driving wheelchairs. There are three ways this system could get into the hands of people who need it. The first would be being purchasing the system themselves and installed on their chairs. In this case it could be released as an open source kit absolving the creators of the burden of responsibility (NOT SURE IF THIS IS TRUE...). There are a number of problems with this method. First of all instead of being held to the level of testing the FDA requires these requirements could be skirted leading to a less safe system. Also with the cost of sensors out of pocket the adoption rate for a system of this kind would be quite low and not as accessible as one would want it to be. Another way would be to have the wheelchair manufactures engineer a device, while the most seamless idea the development cycle would take years to prefect and if done incorrectly could lead to hurt clients and a tarnished image. Finally a startup could develop a system to provide advanced autonomous actions for a wheelchair. Unfortunately unlike self driving cars or small mobile robots the risk of error is higher.

Finally, we must consider risk, both physical and legal, and its role in the adoption of smart wheelchairs. For a person with limited personal mobility, the risks associated with a wheelchair failure are high. If navigation fails, the user may be unable to stop the chair from hitting an obstacle. On robots, this is handled using an emergency stop button. Conversely, many wheelchair users with severe physical disabilities cannot reliably operate such a device. This personal risk leads to a legal risk. If a person is injured using a smart wheelchair, the manufacturer is exposed to legal liability. Since, as in the case of self-driving cars, there is little legal precedent for this type of litigation, the risk exposure is unbounded. This is a significant disincentive for manufacturers to invest in smart wheelchairs.

%% file: conclusions.tex
\section{CONCLUSIONS}

While there has been a lot of research in the area of smart wheelchairs, there are currently very few real-world deployments. We have discussed some of the mismatches between much of the extant research and the realities of using a powered mobility device. As we have tried to move from the laboratory to the real world, we have experienced many of these pitfalls in our own work. Despite the litany of problems highlighted in this paper, we remain extremely optimistic for the future of smart wheelchairs. With a slight change in focus, we strongly believe this technology will be present in everyday life before long.

%The work that has been done with wheelchairs, while great for navigation, has failed to consider the needs and requirements placed on real world systems. Although self driving wheelchairs will eventually be feasible at a large scale, there is research that can be done now to improve the daily lives of people currently in powered wheelchairs that does not require autonomous driving. With an intelligent wheelchair consisting of an IMU, Camera, Odometry Data, a computer, and an internet link. There is much that can be done outside of self driving wheelchairs. Instead of focusing on the long term future goal we should consider what is practical here and now to not only facilitate more advanced wheelchairs in the future, but to provide improved quality of life for people now.

Inspiration can be drawn from the automobile industry. Technologies like lane assist, parking assist, backup cameras, and blind-spot detection have made the driving experience better and safer without full autonomy. As with the automobile industry, autonomous navigation will eventually be developed for smart wheelchairs since it is a desired feature~\cite{gomi:wheelchairdevelopment}. This development relies on many factors such as advancements in sensor technology, rapidly dropping sensor costs~\cite{Edelstein:sensors}, and the support of wheelchair companies. The social and legal precedents currently being set by companies developing commercial autonomous robots will help shape how smart wheelchairs are addressed.

As with anyone with a routine, much of a wheelchair user's time will be spent in only a few locations. Modifying the environment rather than the wheelchair eliminates many of the problems with instrumenting a wheelchair. By developing and deploying Internet of Things (IoT) devices combined with simple on-board sensors, complex features can be realized with relatively little effort.

In the short term, we value focusing on technologies to improve the quality of life of wheelchair users instead of focusing on self-driving wheelchairs. Backup cameras, location-aware chairs that interact with IoT devices, and activity monitors are examples of such enhancements. Technologies like these are more straightforward to implement, rely on the human for robust decision-making, and will help to bridge gap to higher functioning smart wheelchairs.

%instead of focusing on systems that can intervene or completely take over control, we should consider systems that augment the users experience and provide quality of life improvements. The first step is designing systems that cannot harm or increase the risk of harm in a person's life. The simplest way to prevent intelligent wheelchairs from possibly causing harm to the user is to develop passive systems rather than active ones. In a practical sense, none of the new intelligent parts would be allowed to control the chair or make navigation decisions for the user. However, features like reducing interaction complexity, wheelchair control difficulty, and heath monitoring all could have a great impact on a wheelchair users daily life. 

% \todo[inline]{Need something about being more practical about navigation here, but I'm not really sure what. Should be optimistic, since we'll nail it eventually, but it should also say that if we do the early stuff, then we can convince the manufacturers that autonomy is a good thing, so they'll open up their systems, and everything will be easier. Plus, we'll have loose precedent from self-driving cars, and sensors that work well, like radar chips from the auto industry. Maybe there's a subtle way of saying if we're going to be analogizing self-driving cars, then we need to wait until they're a thing first.}